\documentclass[10pt,conference,a4paper]{IEEEtran}
\usepackage{cite}
\usepackage{times}
\usepackage{amsmath,amssymb,amsfonts}
\usepackage{algorithmic}
\usepackage{graphicx}
\usepackage{textcomp}
\usepackage{xcolor}
\usepackage{xspace}
\def\BibTeX{{\rm B\kern-.05em{\sc i\kern-.025em b}\kern-.08em
    T\kern-.1667em\lower.7ex\hbox{E}\kern-.125emX}}

\numberwithin{theorem}{section}

\newcommand{\M}[1]{\mathtt{#1}}
\newcommand{\V}[1]{\mathbf{#1}}

\newcommand{\comment}[1]{}

\newcommand{\etal}{\textit{et al}.}
\newcommand{\ie}{\textit{i}.\textit{e}.}
\def\Hk{Heikkil{\"{a}}\xspace}  
\def\gb{Gr{\"o}bner basis\xspace}

\def\gbs{Gr{\"o}bner bases\xspace}

\def\rs{resultant\xspace}

\newcommand{\PAR}[1]{\vskip4pt \noindent{\bf #1~}}
\def\PfP{$\textrm{P5P}_{\textrm{r}}$\xspace}
\def\PsP{$\textrm{P6Pf}_{\textrm{r}}$\xspace}
\def\PtP{$\textrm{P2P}_{\textrm{r}}$\xspace}

\begin{document}

\title{Computing stable resultant-based minimal solvers by hiding a variable}

\author{\IEEEauthorblockN{Snehal Bhayani$^\textrm{1}$ $\qquad\qquad$
Zuzana Kukelova$^\textrm{2}$ $\qquad\qquad$
Janne \Hk$^\textrm{1}$} 
\IEEEauthorblockA{$^\textrm{1}$Center for Machine Vision and Signal Analysis, University of Oulu, Finland\\
$^\textrm{2}$Visual Recognition Group, Faculty of Electrical Engineering, Czech Technical University in Prague}
}

\maketitle

\begin{abstract}
\noindent Many computer vision applications require robust and efficient estimation of camera geometry. The robust estimation is usually based on solving camera geometry problems from a minimal number of input data measurements, \ie, solving minimal problems, in a RANSAC-style framework. Minimal problems often result in complex systems of polynomial equations. The existing state-of-the-art methods for solving such systems are either based on \gbs and the action matrix method, which have been extensively studied and optimized in the recent years or recently proposed approach based on a \rs computation using an extra variable.

In this paper, we study an interesting alternative \rs-based method for solving sparse systems of polynomial equations by hiding one variable. This approach results in a larger eigenvalue problem than the action matrix and extra variable \rs-based methods; however, it does not need to compute an inverse or elimination of large  matrices that may be numerically unstable. The proposed approach includes several improvements to the standard sparse resultant algorithms, which significantly improves the efficiency and stability of the hidden variable resultant-based solvers as we demonstrate on several interesting computer vision problems. We show that for the studied problems, our sparse resultant based approach leads to more stable solvers than the state-of-the-art \gb as well as existing \rs-based solvers, especially in close to critical configurations. Our new method can be fully automated and incorporated into existing tools for the automatic generation of efficient minimal solvers.
\end{abstract}

\vspace{-0.3cm}
\section{Introduction}
\noindent Computation of the camera geometry is one of the most important tasks in computer vision~\cite{HZ-2003} with many applications e.g., in structure from motion~\cite{Snavely-IJCV-2008}, visual navigation~\cite{DBLP:journals/ram/ScaramuzzaF11}, large scale 3D reconstruction~\cite{DBLP:conf/cvpr/HeinlySDF15} and 
image localization~\cite{Sattler16PAMI}.

The robust estimation of camera geometry is usually based on solving so-called minimal problems~\cite{Nister-5pt-PAMI-2004,Kukelova-ECCV-2008, Kukelova-thesis}, \ie problems that are solved from minimal samples of input data, inside a RANSAC-style framework~\cite{Fischler-Bolles-ACM-1981,Chum-2003, DBLP:journals/pami/RaguramCPMF13}. Since the camera geometry estimation has to be performed many times in RANSAC~\cite{Fischler-Bolles-ACM-1981}, having fast and accurate solvers to minimal problems is of high importance.

Minimal problems usually result in complex systems of polynomial equations in several variables.   A popular approach for solving such problems in computer vision is to design procedures, i.e., specialized polynomial solvers, that can efficiently solve only a special class of systems of equations with a given structure, e.g., systems resulting from the 5-pt relative pose problem~\cite{Nister-5pt-PAMI-2004}.  Such solvers move as much computation as possible from the ``online'' stage of solving equations to an earlier pre-processing ``offline'' stage.

Most of the state-of-the-art minimal solvers are based on \gbs and the action-matrix method~\cite{Cox-Little-etal-05}. The \gb method was popularized in computer vision by Stewenius~\cite{DBLP:phd/basesearch/Stewenius05}. The first efficient \gb solvers were mostly handcrafted~\cite{Stewenius-ISPRS-2006,Stewenius-CVPR-2005} and sometimes very unstable~\cite{stewenius2005hard}. However, in the last 15 years much effort has been put into making the process of constructing the solvers more automatic~\cite{Kukelova-ECCV-2008,larsson2017efficient,larsson2017polynomial} and the solvers stable~\cite{byrod2007improving,byrod2008column} and more efficient~\cite{larsson2017efficient,larsson2017polynomial,larsson2016uncovering,Bujnak-CVPR-2012, DBLP:conf/cvpr/LarssonOAWKP18}. There are now powerful tools available for the automatic generation of efficient \gb solvers \cite{Kukelova-ECCV-2008,larsson2017efficient}.
While the \gb method 
was deeply studied in computer vision and all recently generated \gb solvers are highly optimized in terms of efficiency, less attention has been paid to the \rs-based approach for generating polynomial solvers. Existing resultant-based solvers are mostly handcrafted and tailored to a particular problem, are not exploiting a sparsity of the systems~\cite{Kukelova-thesis} or can not be directly applied to general minimal problems~\cite{DBLP:conf/iccv/Heikkila17}. 
Only recently Bhayani \etal~\cite{bhayani2019sparse} proposed a general sparse \rs-based approach for generating polynomial solvers by augmenting the original system with a polynomial of a special form. Such solvers compute a Schur complement of a special submatrix of the resultant matrix, leading to an compact eigenvalue problem.

Especially in close to degenerate configurations, the matrices that have to be inverted or eliminated in the state-of-the-art \gb~\cite{larsson2017efficient,Kukelova-ECCV-2008} and resultant-based solvers~\cite{bhayani2019sparse} may be close to singular and therefore these solvers may be unstable.

In this paper, we study an alternative \rs-based method for generating efficient minimal solvers where we attempt to improve the solver stability by sacrificing solver speed to a certain extent.
The proposed approach removes the potentially numerically unstable computation of an inverse or Gauss-Jordan elimination of matrices present in state-of-the-art solvers~\cite{larsson2017efficient,bhayani2019sparse} at the cost of a larger eigenvalue problem. Additionally, we propose several improvements to the previously published sparse resultant method~\cite{DBLP:conf/iccv/Heikkila17}.
%
The new approach leads to more stable solvers as compared to ones based \gb as well as the recent \rs-based solvers~\cite{bhayani2019sparse} especially in close-to-degenerate configurations  as we demonstrate on three interesting computer vision problems while maintaining a comparable solver speed. Our new method can be fully automated and incorporated in existing tools for automatic generation of efficient minimal solvers~\cite{Kukelova-ECCV-2008,larsson2017efficient,larsson2017polynomial} and as such applied to a large variety of minimal problems.  
Specifically our contributions include:
\vspace{-0.1cm}
 \begin{itemize}
 \itemsep-0.1em
    \item Several improvements to hidden variable sparse \rs based algorithms~\cite{DBLP:journals/jacm/CannyE00,DBLP:conf/iccv/Heikkila17} that can generate smaller and stable solvers for general polynomial systems.
    \item Replacing a potentially numerically unstable 
     computation of a matrix inverse present in the state-of-the-art \gb~\cite{larsson2017efficient,Kukelova-ECCV-2008} and resultant-based solvers~\cite{bhayani2019sparse}
     with a larger eigenvalue problem computation that is usually numerically more stable.
     \item Demonstrating improved stability, especially in close-to-degenerate configurations, as shown on three interesting absolute pose estimation problems~\cite{haner2015absolute} as compared to state-of-the-art solvers.
 \end{itemize}

\section{Theoretical Background and Related work}
\noindent In this paper we use the notation and basic concepts from algebraic geometry book from Cox \etal \cite{Cox-Little-etal-05} and consider a system of $m$ polynomial equations, 
\begin{eqnarray}\label{eq:system}
\lbrace f_1(x_1,...,x_n)=0,...,f_m(x_1,...,x_n)=0 \rbrace
\end{eqnarray}
in $n$ unknowns $X = \lbrace x_1,...,x_n\rbrace$, where $m \geq n$ and with a finite number of solutions. Our objective is to compute the solutions to this system.

\subsection{Sparse Resultants and eigenvalue problems} 
\noindent For the input polynomial system in \eqref{eq:system} with $m = n+1$, a resultant is defined as an irreducible polynomial constraining its coefficients to have non-trivial solutions. For a more formal theory of resultants and their properties, we refer to Cox \etal~\cite{Cox-Little-etal-05}.

We note that standard resultants are defined for a polynomial system~\eqref{eq:system} with $m = n+1$, where the coefficients are also considered as variables~\cite{Cox-Little-etal-05}.  This originates from the fact that resultants were initially developed to determine whether a system of $n+1$ polynomial equations in $n$ unknowns has a common root or not. Denoting a coefficient of the monomial $x^{\alpha}$ in $i^{th}$ polynomial as $u_{i,\alpha}$ we have the resultant, $Res([u_{i,\alpha}])$ as a polynomial in $u_{i, \alpha}$ as variables.

Using these notations as well as terminology, the basic idea for a resultant-based method is to expand the polynomials $f_{1},...,f_{n+1}$ to a set of linearly independent polynomials which can be written in a matrix form 
\begin{equation} \label{eq:spresconstraint}
    \M{M}([u_{i, \alpha}]) \V{x},
\end{equation}
where $\M{M}([u_{i, \alpha}])$ has to be a square matrix that is full rank for generic values of $u_{i,\alpha}$, i.e. $\det \M{M}([u_{i, \alpha}]) \neq 0$. The determinant of the matrix $\M{M}([u_{i, \alpha}])$ is a non-trivial multiple of the resultant $Res([u_{i, \alpha}])$~\cite{Cox-Little-etal-05}. Thus determinant $\det \M{M}([u_{i, \alpha}])$ must vanish, if the resultant vanishes, \ie 
$Res([u_{i, \alpha}]) = 0 \implies \det \M{M}([u_{i, \alpha}]) = 0$. It is known that $Res([u_{i, \alpha}])$ vanishes iff the system $f_{1},...,f_{n+1}$ has a solution~\cite{Cox-Little-etal-05}. 
This gives us a necessary condition for the existence of roots of $f_{1},...,f_{n+1}$. 
Hence equation $\det \M{M}([u_{i, \alpha}]) = 0$ gives us those sets of $u_{i, \alpha}$ such that $f_{1},...,f_{n+1}$ have a common root. In this way traditional resultants can be used to solve a polynomial system~\eqref{eq:system} where $m=n$ either by ``hiding'' one unknown in the coefficient field or by adding an additional polynomial to the original system, which is known as the u-resultant approach~\cite{Cox-Little-etal-05}. 

The u-resultant approach has inspired recent resultant-based method~\cite{bhayani2019sparse} for generating efficient polynomial solvers for systems~\eqref{eq:system}. In~\cite{bhayani2019sparse}  the idea is to \textit{add a new polynomial} $f_{m+1} = x_{i} - \lambda$ by introducing a new variable $\lambda$. The resultant $Res([u_{i, \alpha}], \lambda)$ is then computed for the augmented system by considering $\lambda$ as a constant. The corresponding matrix $\M{M}([u_{i, \alpha}])$ in \eqref{eq:spresconstraint} is linear in $\lambda$ and for a cleverly chosen submatrix, its Schur complement gives a compact eigenvalue formulation whose eigenvectors provide solutions to $x_{1},...,x_{n}$.

In this paper we explore an alternative approach for solving~\eqref{eq:system} using resultants by \textit{hiding a variable} or in other words considering one of the existing variables (say $x_{n}$) as constant. In this way the resultant $Res([u_{i, \alpha}], x_{n})$ becomes a function of $u_{i, \alpha}$ and $x_{n}$. Specifically our proposed algorithm attempts to expand a polynomial system to a linearly independent set of polynomials that can be re-written in a matrix form as
\begin{equation} \label{eq:spresconstraint2}
    \M{M^\prime}([u_{i, \alpha}],x_{n}) \V{x^\prime} = 0, 
\end{equation}
where $\M{M^\prime}([u_{i, \alpha}],x_{n})$ is a square matrix whose elements are polynomials in $x_{n}$ and coefficients $u_{i, \alpha}$ and $\V{x}^\prime$ is the vector of monomials in $x_{1},...,x_{n-1}$.
For simplicity we will denote the matrix $\M{M^\prime}([u_{i, \alpha}],x_{n})$  as $\M{M^\prime}(x_{n})$ in the rest. 
Similar to the state-of-the-art methods~\cite{DBLP:journals/jacm/CannyE00,DBLP:conf/iccv/Heikkila17}, we actually estimate a multiple of the resultant via the determinant of the matrix $\M{M^\prime}(x_{n})$ in~\eqref{eq:spresconstraint2}. This resultant is known as a hidden variable resultant and it is a polynomial in $x_n$ whose roots are the $x_{n}$-coordinates of the solutions of the system of polynomial equations. For theoretical details and proofs see \cite{Cox-Little-etal-05}.

One way to solve the original system of polynomial equations is to compute the roots of the polynomial $\det \M{M^\prime}(x_{n}) =0$ and then, after substituting solutions for $x_{n}$ to~\eqref{eq:spresconstraint2}, extract the solutions to the remaining variables from the right eigenvectors of the matrix $\M{M^\prime}(x_{n})$. Unfortunately, computing a determinant of a large polynomial matrix $\M{M^\prime}(x_{n})$ may be numerically unstable. Therefore, this problem is usually transformed to a polynomial eigenvalue problem (PEP)~\cite{Cox-IVA-2015}.

The matrix equation~\eqref{eq:spresconstraint2} can be re-written in a PEP form  
\begin{equation}\label{eq:pepformulation}
(\M{M}_{0} + \M{M}_{1} \ x_{n}+...+ \M{M}_{l} \ x^{l}_{n})\V{x^\prime} = \V{0},
\end{equation}
where $l$ is the degree of the matrix $\M{M^\prime}(x_{n})$ in the hidden variable $x_n$ and matrices $\M{M}_{0},...,\M{M}_{l}$ are matrices that depend only on the coefficients $u_{i, \alpha}$ of the original system of polynomials. PEP~\eqref{eq:pepformulation} can be easily converted to a generalized eigenvalue problem (GEP)
\begin{equation}\label{eq:GEP}
\M{A} \V{y} =  x_{n} \M{B} \V{y},
\end{equation}
and solved using standard efficient eigenvalue algorithms. Basically, the eigenvalues give us the solution to $x_{n}$ and the rest of the variables can be solved from the corresponding eigenvectors, $\V{y}$, ~\cite{Cox-Little-etal-05}. 
We note that this transformation to a GEP~\eqref{eq:GEP} is a relaxation of the original problem of finding the solutions to our input system. First of all, when we are searching for the eigenvalues and the eigenvectors of PEP~\eqref{eq:pepformulation} we are not considering monomial dependencies induced by the monomial vector $\V{x^\prime}$. Actually, we are linearizing the original system equations~\eqref{eq:system} and therefore beside the ``correct`` eigenvectors where the dependencies are satisfied we also get eigenvectors where these dependencies do not hold. Second, transforming PEP~\eqref{eq:pepformulation} to a GEP~\eqref{eq:GEP} usually introduces additional parasitic (zero) eigenvalues.
 
The hidden variable approach was used to solve various minimal problems in computer vision. Kukelova \etal \cite{Kukelova-PolyEig-PAMI-2012, Kukelova-thesis} and Hartley \etal~\cite{Hartley-PAMI-2012} used the hidden variable approach to solve important problems of estimating relative pose of two calibrated cameras (5-pt relative pose problem), and cameras with unknown focal lengths (6-pt relative pose problems). 
To improve the numerical stability of the computation of coefficients and the roots of the polynomial determinant, the authors of~\cite{Hartley-PAMI-2012} suggested to use  several numerical techniques including quotient-free Gaussian elimination, Levinson-Durbin iteration and root polishing. All minimal problems presented in \cite{Kukelova-PolyEig-PAMI-2012, Kukelova-thesis, Hartley-PAMI-2012} were simple in the sense that after hiding one variable it was directly possible to rewrite the original systems of polynomial equations in the form ~\eqref{eq:spresconstraint2}, with square matrix $\M{M^\prime}(x_{n})$. Unfortunately this is not usually the case, and for more complicated problems, we usually need to generate an extended set of linearly independent polynomials to obtain a square matrix in~\eqref{eq:spresconstraint2}.

In~\cite{DBLP:conf/wacv/KastenGB19}, authors used Dixon resultant matrix to create such extended set of linearly independent polynomials to solve the six point incremental camera pose problem. However, complicated symbolic expressions being used to compute this resultant in runtime leads to a slow solver.

Kukelova~\cite{Kukelova-thesis} proposed a method for generating extended sets of linearly independent polynomials~\eqref{eq:spresconstraint2}. This method was a modification of the Macaulay’s method for computing resultants. Unfortunately, the proposed method is not general and it does not work for all systems of polynomial equations. Macaulay’s method for computing resultants was designed for dense systems with generic coefficients. For sparse systems, which are usually common in computer vision applications, the Macaulay’s method and even the modification proposed in~\cite{Kukelova-thesis} may generate linearly dependent equations and therefore not square matrix in~\eqref{eq:spresconstraint2}. Moreover, for some systems the proposed method was generating unnecessarily many polynomials.

For sparse systems it is possible to obtain a more compact resultant using specialized algorithms. Such resultants are 
commonly referred to as the \textit{Sparse Resultants}. A compact resultant would mostly lead to a more compact matrix $\M{M^\prime}(x_{n})$ and hence a smaller eigendecomposition problem. 

Emiris \etal \cite{DBLP:conf/issac/EmirisC93}, \cite{DBLP:journals/jacm/CannyE00} proposed a generic algorithm using mixed-subdivision of polytopes to estimate the matrix based on the sparse resultants. In~\cite{DBLP:journals/corr/abs-1201-5810} Emiris have applied this method to the 5-point relative pose problem, however the final solver isn't particularly efficient. Recently Bhayani \etal~\cite{bhayani2019sparse} proposed a sparse resultant based approach by introducing an extra polynomial of a special form and computing Schur complement to obtain a small eigendecomposition problem. However, the resulting solvers can become unstable if the matrix that is to be inverted becomes close to singular. This happens especially in close-to-degenerate configurations. 

Another sparse hidden variable resultant based approach has been proposed by Heikkil{\"{a}} in \cite{DBLP:conf/iccv/Heikkila17}. This approach tests and extracts smaller $\M{M^\prime}(x_n)$  as compared to the ones constructed by the Canny-Emiris algorithm  \cite{DBLP:journals/jacm/CannyE00}. The proposed algorithm was tested on a problem of planar self-calibration. The structure of the matrix $\M{M^\prime}(x_n)$~\eqref{eq:spresconstraint2} is here computed only once in the pre-processing step. The matrix contains only monomial multiples of the input polynomials, \ie, its elements are just shifted coefficients of the original polynomials. Hence, the final online solver can simply feed in actual values based on real data and estimate solutions based on eigenvalues and eigenvectors computed after transforming~\eqref{eq:spresconstraint2} to a GEP~\eqref{eq:GEP}.

Our proposed approach builds on top of the method presented in \cite{DBLP:conf/iccv/Heikkila17}. Our contribution is a set of improvements for estimating a more compact monomial basis vector $\V{x^\prime}$ or a more stable matrix $\M{M^\prime}(x_n)$ than the one obtained by the algorithm from \cite{DBLP:conf/iccv/Heikkila17}. Additionally, our approach does not have to compute a Schur complement as is needed by the state-of-the-art \rs-based approach~\cite{bhayani2019sparse}, which helps to improve the solver stability. We briefly describe our algorithm in Section \ref{sec:improvements} and then proceed to highlight existing drawbacks and our proposed improvements.

\subsection{Important features of our proposed algorithm}
\noindent Before we list proposed improvements, 
we would like to highlight important differences between the existing state-of-the-art methods based on \gb as well as resultants for solving systems of polynomial equations~\eqref{eq:system}. We start with mentioning how the methods transform the original problem to an eigendecomposition problem.

Let $k$ be the actual number of solutions to the problem. The \gb method~\cite{larsson2017efficient,Kukelova-ECCV-2008} transforms the problem to that of finding the eigenvalues and eigenvectors of a $k \times k$ matrix, known as the action matrix $\M{M}_f$. Whereas the recently published \rs-based method~\cite{bhayani2019sparse} solves eigenvalues and eigenvectors of a Schur complement of the resultant matrix. The coefficients of such matrices are polynomial combinations of the coefficients of the input polynomials~\eqref{eq:system}. To obtain these coefficients, either Gauss-Jordan elimination needs to be performed on a \textit{special elimination template matrix}~\cite{larsson2017efficient,Kukelova-ECCV-2008}  or matrix inverse and subsequent multiplication has to be performed on submatrices of the resultant matrix~\cite{bhayani2019sparse}. For more complicated systems, such matrices can be very large and sometimes ill-conditioned, leading to numerically unstable solvers. Such ill-conditioned matrices appear especially in close-to-degenerate configurations, e.g., close-to-planar scenes or special type of motions and point configurations, that in some applications may be quite common.

On the other hand, our proposed \rs-based method is a relaxation of the original problem in the sense that it linearizes the original system of polynomial equations~\eqref{eq:system}. Especially, the approach of hiding a variable, transforms the original problem to the problem of finding the eigenvalues and the eigenvectors of matrices $\M{A}$ and $\M{B}$~\eqref{eq:GEP}. In general the sizes of these matrices are larger than $k$ and not all eigenvectors satisfy monomial dependencies induced by the monomial vector $\V{x}^\prime$~\eqref{eq:spresconstraint2}. However, among the eigenvalues and eigenvectors of~\eqref{eq:GEP} there are all solutions to the original system~\eqref{eq:system}. The important difference is that the matrices $\M{A}$ and $\M{B}$ in~\eqref{eq:GEP} contain only the coefficients of the original equations~\eqref{eq:system}. Therefore, once we find the structure of these matrices (which monomial multiples of original equations they contain), we have these matrices ``for free".

We believe that for some problems, eigendecomposition of matrices $\M{A}$ and $\M{B}$ in~\eqref{eq:GEP} may be more efficient and especially numerically more stable than the above mentioned matrix operations, followed by eigendecomposition of either the action matrix in \gb solvers~\cite{larsson2017efficient,Kukelova-ECCV-2008} or Schur complement of the resultant in state-of-the-art extra variable based \rs solvers~\cite{bhayani2019sparse}.

\section{Improving existing sparse resultant based methods }\label{sec:improvements}
\noindent In the literature, sparse resultants are computed by employing the theory of convex polytopes. So, we first define relevant terms, that are common to the existing state-of-the-art sparse resultant algorithms\cite{DBLP:journals/jacm/CannyE00,DBLP:conf/iccv/Heikkila17, bhayani2019sparse}, and then list the drawbacks of existing methods followed by our proposed improvements.

\subsection{Monomial basis selection with convex polytopes}
\noindent A Newton polytope of a polynomial $NP(f)$ is defined as a convex hull of the exponent vectors of the monomials occurring in the polynomial (also known as the support of the polynomial). Hence, we have $NP(f_{i}) = \text{Conv}(A_{i})$ where $A_{i} = \lbrace \alpha | \alpha \in \mathbb{Z}_{n} \rbrace$ is the set of all integer vectors that are exponents of monomials with non-zero coefficients in $f_{i}$. A Minkowski sum of any two convex polytopes $P_1, P_2$ is defined as $P_1+P_2 = \lbrace p_1 + p_2 \ | \ \forall p_1 \in P_1, p_2 \in P_2 \rbrace$. An extensive treatment of polytopes can be found from \cite{Cox-Little-etal-05}. 

The basic idea in the \textit{Canny-Emiris algorithm} \cite{DBLP:journals/jacm/CannyE00} is to calculate the Minkowski sum of the Newton polytopes of all input polynomials, $ Q=  \Sigma_{i} NP(f_{i})$. The set of integer points in the interior of $Q$ defined as $B = \mathbb{Z}_{n-1} \cap (Q + \delta)$, where $\delta$ is a small random displacement vector, can provide a monomial basis $\V{x^\prime}$ satisfying the constraint~\eqref{eq:spresconstraint2}. 

Heikkil{\"{a}} \cite{DBLP:conf/iccv/Heikkila17} and Bhayani \etal~\cite{bhayani2019sparse} use the same principle of extracting the basis monomials from the Minkowski sum of the Newton polytopes, but instead of summing all the polytopes, they summed only subsets of the polytopes with different combinations. 

The main steps of \textit{Heikkil{\"{a}}'s algorithm} are:
 \vspace{-0.1cm} 
\begin{enumerate}
  \itemsep0em 
    \item Given $n$ polynomials, $f_{1},...,f_{n}$ in $n$ unknowns, $x_{1},...,x_{n}$ hide one variable, say $x_{n}$, to the coefficient field. Determine the Newton polytopes $NP_{i}$ for each new polynomial in the $n-1$ dimensional space.
    \item Calculate the Minkowski sum of all combinations $j$ of the Newton polytopes $\lbrace Q_{j}=  \Sigma_{i(j)} NP(f_{i})\rbrace$.
    \item Create a set $\lbrace B_{j,k} = \mathbb{Z}_{n-1} \cap ( Q_{j} + \delta_{k})\rbrace$ by using all possible displacement vectors $\lbrace \delta_k \rbrace$ with elements in $\lbrace -\epsilon,0,\epsilon \rbrace$, where $\epsilon$ is a small positive constant. The elements of $B_{j,k}$ are exponential vectors that form a prospective monomial basis.
    \item For every $B_{j,k}$ find a set of monomials $\lbrace T_{j,k}\rbrace$ that are used to multiply the original polynomials to get $|T_{j,k}|$ linearly independent polynomials within the basis $B_{j,k}$. Accept the basis only if every original polynomial contributes to the new set of polynomials and $|T_{j,k}| \geq |B_{j,k}|$ to guarantee that $\M{M^\prime}(x_{n})$ has at least as many rows as columns.
    \item Finally, select the smallest basis $B_{j,k}$ that fulfills the previous conditions and assign it to $\V{x^\prime}$.
\end{enumerate}

\subsection{Drawbacks}
\noindent Next we list shortcomings of the previous methods~\cite{DBLP:journals/jacm/CannyE00,DBLP:conf/iccv/Heikkila17,bhayani2019sparse}:
\vspace{-0.3cm}
\begin{enumerate} 
 \itemsep0em 
    \item The approaches in~\cite{DBLP:journals/jacm/CannyE00,DBLP:conf/iccv/Heikkila17} assume that the number of polynomials to be solved is exactly the same as the number of unknowns. However, many minimal problems in computer vision have actually more equations than unknowns. Due to the intrinsic properties of the Canny-Emiris algorithm it does not allow to take into account the additional equations, while Heikkil{\"{a}}'s method does not pose any restrictions to the number of equations, and it can be easily extended to cover such problems. 
    \item Heikkil{\"{a}}'s algorithm~\cite{DBLP:conf/iccv/Heikkila17} can result in a monomial basis such that the matrix $\M{M^\prime}(x_{n})$ in ~\eqref{eq:spresconstraint2} is rank deficient and hence leads to unstable or incorrect solvers.
    \item Heikkil{\"{a}}'s algorithm~\cite{DBLP:conf/iccv/Heikkila17} is used with the PEP formulation, where the matrix $\M{M^\prime}(x_{n})$ is converted to $\M{A}$ and $\M{B}$ matrices of the GEP problem~ \eqref{eq:GEP}. Such conversion leads to large and sparse matrices, which introduces parasitic eigenvalues that are either $0$ or $\infty$. This results in a computationally inefficient solver.
    \item The approach by Bhayani \etal~\cite{bhayani2019sparse} involves computation of matrix inverse which may lead to unstable solvers for complex polynomial systems.
\end{enumerate}

\subsection{Proposed improvements and extensions}
\label{subsec:zeigrem}  
\noindent We now propose certain improvements by extending Heikkil{\"{a}}'s algorithm~\cite{DBLP:conf/iccv/Heikkila17} to resolve these drawbacks.

\vspace{-0.1cm}
\PAR{Additional equations}
\newline
We relax the requirement of having the same number of equations and unknowns, and assume that there are $m \geq n$ polynomial equations with $n$ unknowns~\eqref{eq:system} in Step 1 of the algorithm. 
We also perform an exhaustive search across all polynomial combinations and variables by hiding each variable $x_i$ at a time. This usually reduces the monomial basis size leading to a smaller matrix $\M{M^\prime}(x_{n})$ than the matrix generated by  Heikkil{\"{a}}'s algorithm~\cite{DBLP:conf/iccv/Heikkila17}.

\vspace{-0.1cm}    
\PAR{Rank constraint for $\M{M^\prime}(x_{n})$}
\newline
The problem of rank deficiency is resolved by testing for rank of the matrix $\M{M^\prime}(x_{n})$ for every prospective monomial basis $B_{j,k}$ in Step 4. This guarantees that the eigenvalues and eigenvectors of GEP formulation for this matrix~\eqref{eq:GEP} provide solutions to the original polynomial set.
    
\vspace{-0.1cm}    
\PAR{Removal of parasitic eigenvalues}          
\newline
We know that a GEP formulation~\eqref{eq:GEP} for many minimal problems in computer vision has parasitic zero (or $\infty$) eigenvalues due to zero columns in $\M{A}$(or $\M{B}$). Here we outline a simple process for removal of such eigenvalues by eliminating those row-column pairs.

If $\M{A}$ and $\M{B}$ are $k \times k $ matrices, the idea here is to choose a zero column (say at index $j$) in $\M{A}$ that corresponds to a column in $\M{B}$ with at most one non-zero entry. Assuming the row $i$ in $\M{B}$ to have a non-zero value, this row-column pair can be removed while preserving the non-zero eigenvalues. This removal gives us matrices, $\M{A_{r}}$ and $\M{B_{r}}$ for which the same formulation in \eqref{eq:GEP} holds. \begin{equation}\label{eq:modifiedGEP} 
    \M{A_{r}}\V{y_{r}} =  x_{n}\M{B_{r}}\V{y_{r}}. 
\end{equation} Here $\V{y_{r}}$ denotes the reduced eigenvector after removing the element in row $j$. We note that even though the reduced vector $\V{y_r} \neq \V{y}$,  we can still extract the correct values to the rest of the unknowns. This step can be performed again on the reduced pair of matrices leading to further reduction until we can no longer find a zero column in $\M{A}$ or a corresponding column with not more than one non-zero entry in $\M{B}$. This simple idea for removing zero eigenvalues was proposed already in~\cite{Kukelova-PolyEig-PAMI-2012, Kukelova-thesis}.

In case that there are columns in $\M{B}$ with more than one non-zero value, corresponding to zero columns in $\M{A}$, the method from~\cite{Kukelova-PolyEig-PAMI-2012, Kukelova-thesis} does not work. So we perform a specialized variant of row elimination to transform such matrices. After this transformation, we have more row-column pairs that satisfy the criterion mentioned above. We outline this approach by assuming that column $j$ is a zero column in $\M{A}$ and the corresponding column in $\M{B}$ has two non-zero values, in $i_{1}$ and $i_{2}$ positions. Hence, \begin{small}
    \begin{equation}
        \M{A} = \begin{bmatrix}
        \M{a}^{\prime}_{k \times j-1} & \M{0}_{k \times 1} & \M{a}^{\prime \prime \prime}_{m \times (k-j)}
        \end{bmatrix}, 
    \end{equation} \begin{equation}
    \M{B} = \left[ \! \begin{array}{ccc}
            \M{b}^{\prime}_{k \times j-1}  \! & \! \M{b}^{\prime \prime}_{k \times 1}   \! & \! \M{b}^{\prime \prime \prime}_{k \times (k-j)} \\
    \end{array} \! \right]
    \end{equation}
    \end{small} where $\M{b}^{\prime \prime} = \left[ 0 \ \cdots \ b_1 \ \cdots \ 0 \ \cdots \ b_2 \ \cdots \ 0 \right]^T$ is column vector with non-zero values $b_1$ and $b_2$ in positions $i_1$ and $i_2$ respectively. Then we can easily find an special matrix, $\M{G}_{k \times k}$ that pre-multiplies $\M{A}$ and $\M{B}$ such that it performs elementary row operation on the $i_2$th row and transforms its value, $b_2$ in $j$th column to zero. An example of $\M{G}$ that Gauss-eliminates the $i_2$th row containing $b_2$ in previous example is \begin{small}
    \begin{equation}
    \M{G} = \left[
    \begin{array}{c|cccccccc}
            & 1 & 2 & i_1 & . & i_2 & . & .   \\
            \hline
            1 & 1 & . & . & . & . & . & .  \\
            2 & . & 1 & . & . & . & . & . \\
            3 & . & . & 1 & . & . & . & . \\
            . & . & . & . & 1 & . & . & . \\
            i_2 & . & . & -b_2/b_1 & . & 1 & . & .  \\
             . & . & . & . & . & . & 1 & . \\
            . & . & . & . & . & . & . & 1 \\
 \end{array}
    \right]_{k \times k}. 
 \end{equation}
\end{small} The matrix is the same as that of an identity matrix except that its $i_2$th row contains a non-zero value at $i_1$th column. We can repeat this step to transform as many columns as possible in $\M{B}$ to have only one non-zero entry for each zero column of $\M{A}$. If we perform this step $l$ times, we have $l$ such special matrices, $\M{G_1},...,\M{G_l}$, and pre-multiplying $\M{A}$ and $\M{B}$ with them gives us $l$ prospective row-column pairs that satisfy the condition laid out for row-column removal. Removing these pairs gives us reduced matrices and they can be tested for removal of row-column pairs. This step can be repeatedly performed and we only record the row operations performed by each of the special matrices, $\M{G_{1}}...\M{G_{l}}$. Basically, these operations when performed on any matrix pair satisfying GEP~\eqref{eq:GEP}, it gets transformed to $\M{A^\dagger} \V{y^\dagger} =  x_{n} \M{B^\dagger} \V{y^\dagger}$, where $\M{A^\dagger}$ and $\M{B^\dagger}$ are the reduced matrices after removing row-columns from $\M{A}$ and $\M{B}$. By swapping the position of $\M{A}$ and $\M{B}$ in \eqref{eq:GEP} we can remove $\infty$ parasitic eigenvalues by reapplying this exact step. 
We note that all of these improvement steps are performed during the offline stage, for a given problem. The output of the offline steps are simple row operations (recorded as a template by $\M{G}$) and templates for $\M{A}$ and $\M{B}$ matrices in GEP~\eqref{eq:GEP}. At runtime, the recorded row operations are performed on input $\M{A}$ and $\M{B}$ matrices and converted to a reduced GEP containing $\M{A^\dagger}$ and $\M{B^\dagger}$ whose eigendecomposition gives us solutions. 

\section{Absolute pose estimation for flat refractive surfaces}
\noindent We consider three absolute pose estimation problems for flat refractive surfaces~\cite{haner2015absolute} in order to test the improvements proposed in the previous section. We compare the new resultant-based solver with the original solvers for these three problems~\cite{haner2015absolute}, the \gb solvers generated by the state-of-the-art automatic generator~\cite{larsson2017efficient}, improved \gb solvers based on heuristic presented in~\cite{DBLP:conf/cvpr/LarssonOAWKP18}, the Canny-Emiris algorithm~\cite{DBLP:journals/jacm/CannyE00}, the hidden variable approach by \Hk~\cite{DBLP:conf/iccv/Heikkila17} and the state-of-the art resultant-based solvers~\cite{bhayani2019sparse}. We do not compare our solutions to the Gr\"{o}bner fan solvers presented in ~\cite{DBLP:conf/cvpr/LarssonOAWKP18} since for two of three considered problems we were not able to generate solvers using this method in reasonable time.   

The scene geometry of the considered problems involves the ray connecting the camera center $\V{C}$ and the image point, $\vec{\V{u}}$, and the refracted ray through the medium $\vec{\V{v}}$ which passes through the scene point $\V{X}$. Based on the Snell's law~\cite{haner2015absolute}, 
the refracted ray $\V{v}$, the incident ray $\V{u}$ and the refractive plane normal $\vec{\V{n}}$ are coplanar and hence for each scene-image point correspondence, we would have one polynomial constraint. For more details on the scene geometry as well as the problem formulations, we refer to \cite{haner2015absolute} and list the basic details of the three minimal problems in the next section.
\subsection{Five and six point absolute pose problems}\label{subsec:5pt}
For a calibrated camera, there are $5$ degrees of freedom ($3$ for rotation $\M{R}$, $2$ for translation $\V{t}$). As each point correspondence leads to one coplanarity constraint, we need $5$ point correspondences to solve for $\M{R}$ and $\V{t}$. The problem is denoted as \PfP. For a with unknown focal length, there would be $1$ more degree of freedom, and hence leads to a six point absolute pose problem, denoted as \PsP.

\subsection{Absolute pose with known rotation axis}
This is a reduced problem with the assumption that the rotation axis is known. Assuming that the known axis coincides the y-axis, we have only $1$ rotational degrees of freedom. 
Thus as there are $4$ degrees of freedom in all ($3$ for the translation vector and $1$ for the rotation around the known axis), and we need $2$ point correspondences.
The input to the solver is $6$ polynomial equations, $4$ from the Snells law and $2$ from coplanarity constraint. We denote this problem as \PtP.
\begin{figure*}[t]
\centering
\includegraphics[width=0.24\linewidth]{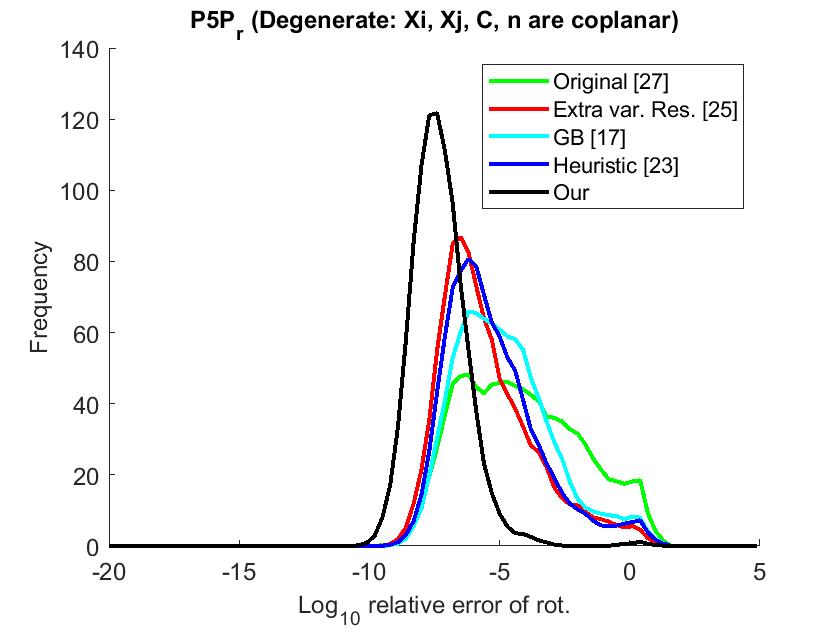}
\includegraphics[width=0.24\linewidth]{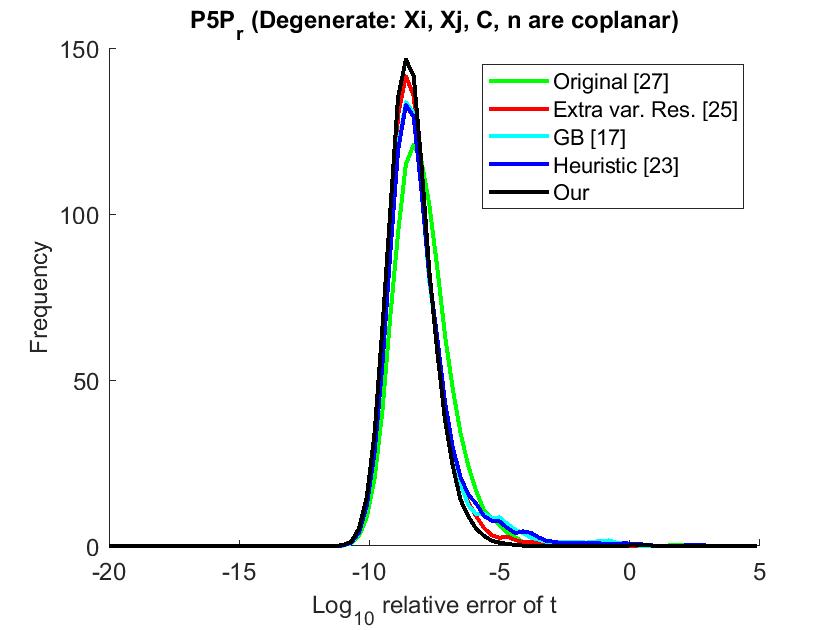}
\includegraphics[width=0.24\linewidth]{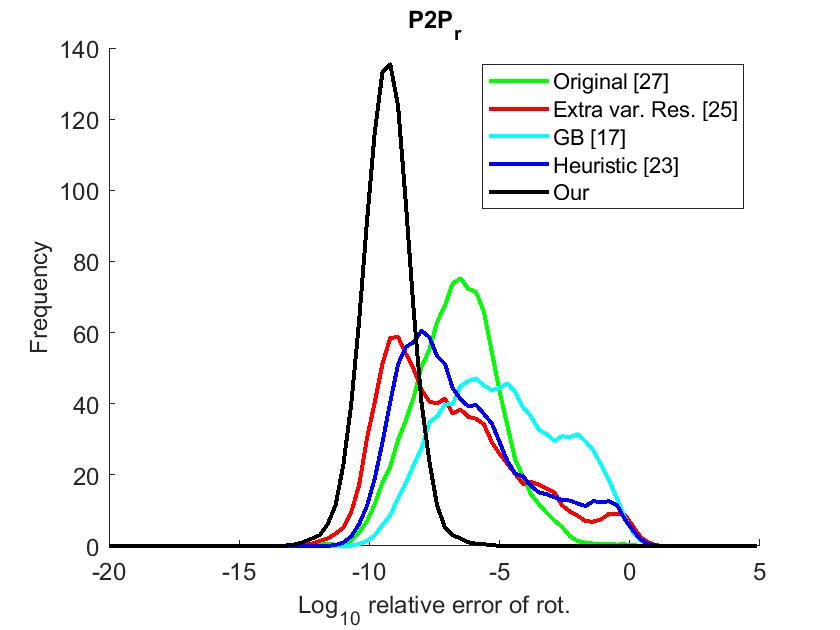}
\includegraphics[width=0.24\linewidth]{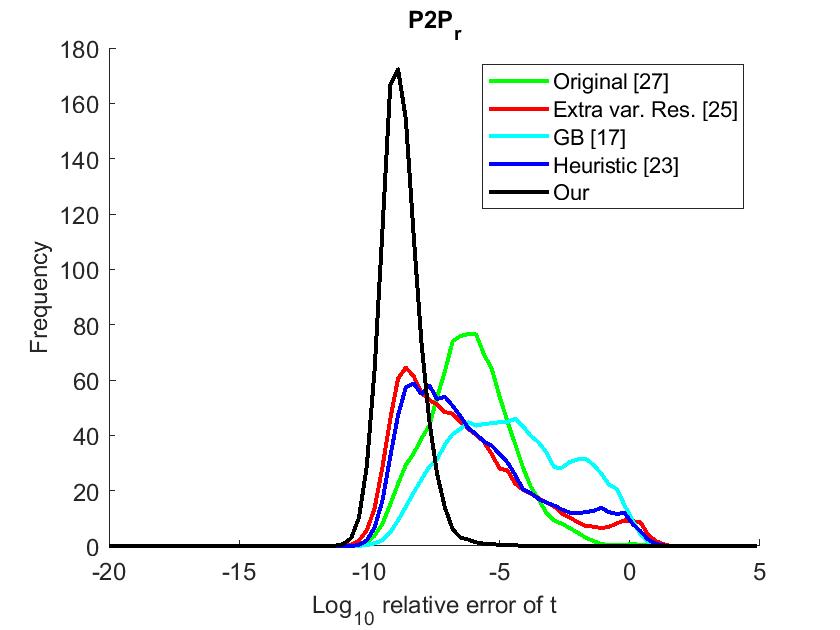}\\
\includegraphics[width=0.24\linewidth]{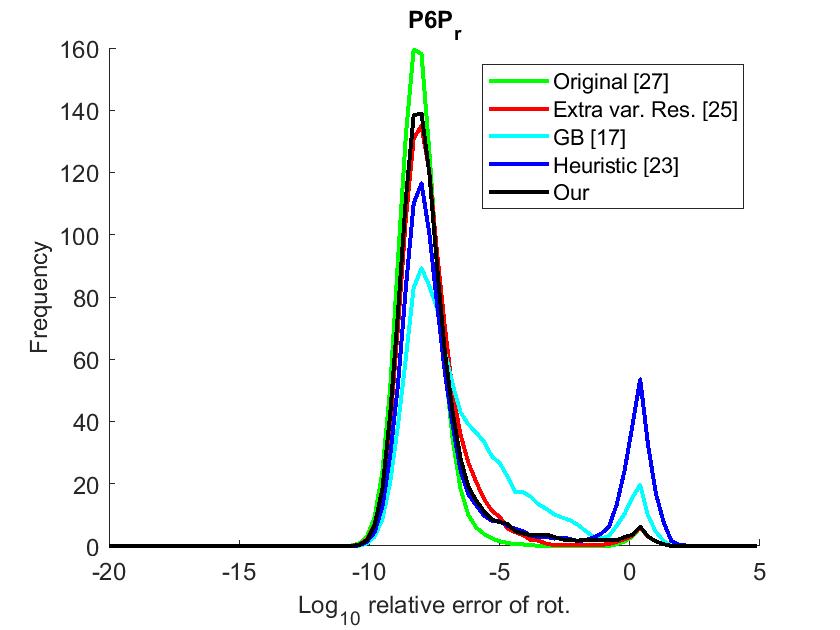}
\includegraphics[width=0.24\linewidth]{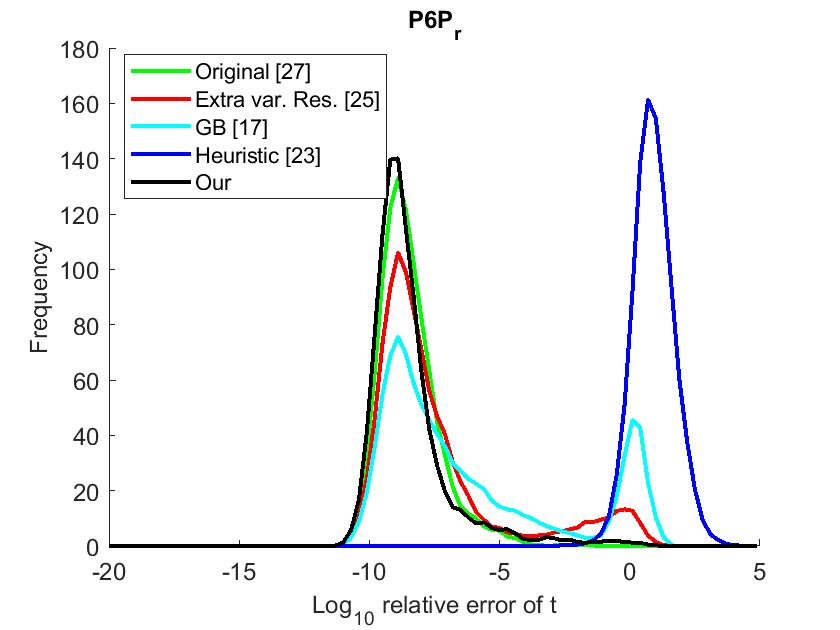}
\includegraphics[width=0.24\linewidth]{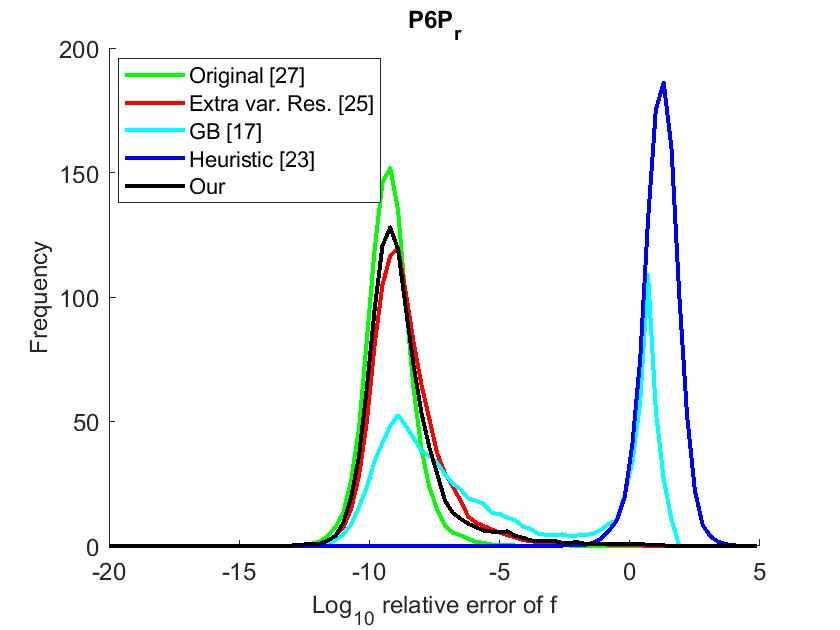}
 \caption{Top: Relative errors of (a) rotation and (b) translation for the \PfP problem, (c) rotation and (d) translation for the \PtP problem measured w.r.t. ground truth on 1K synthetic scenes. Bottom: Relative errors of (e) rotation, (f) translation and (g) focal length for the \PsP problem measured w.r.t. ground truth on 1K synthetic scenes. }
\label{fig:randscenehistograms}
\vspace{-0.1cm}
\end{figure*}
\begin{figure*}[t]
\centering
\includegraphics[width=0.24\linewidth]{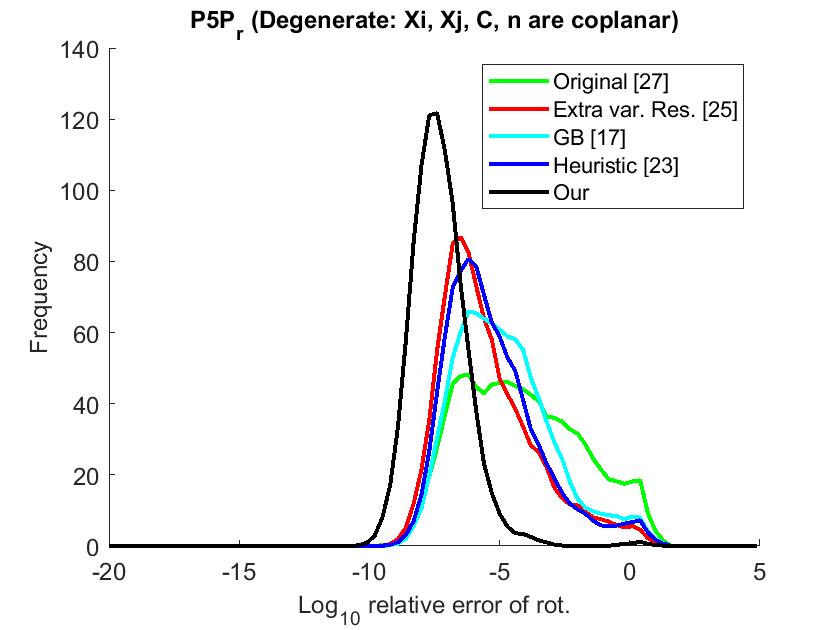}
\includegraphics[width=0.24\linewidth]{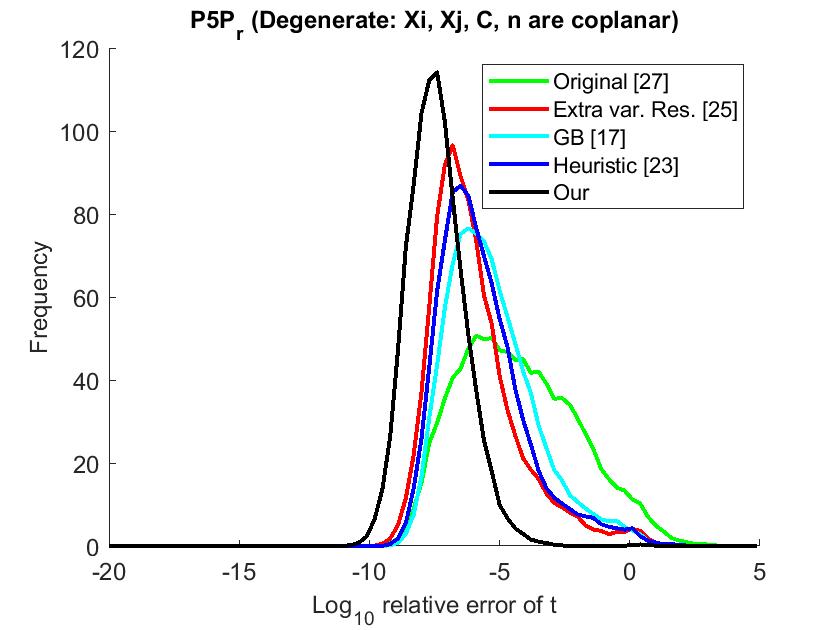}
\includegraphics[width=0.24\linewidth]{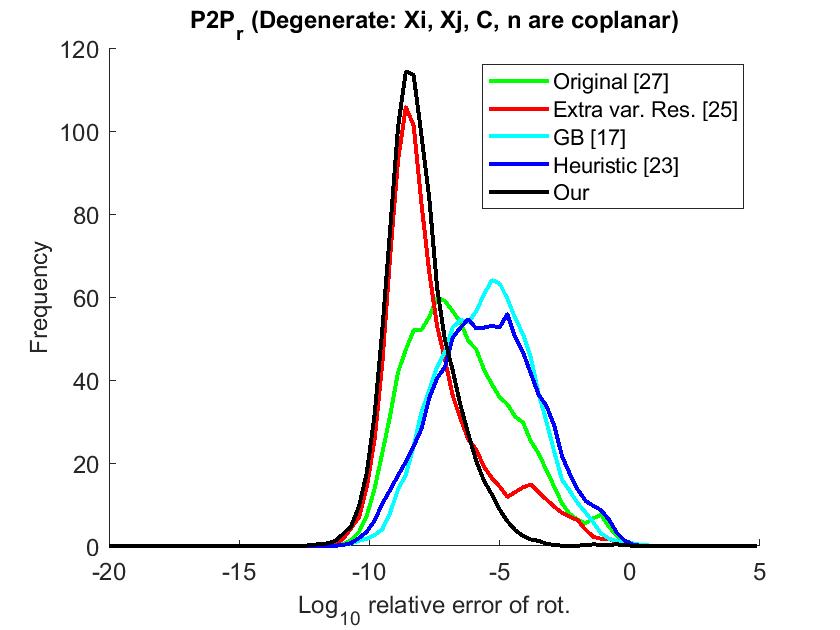}
\includegraphics[width=0.24\linewidth]{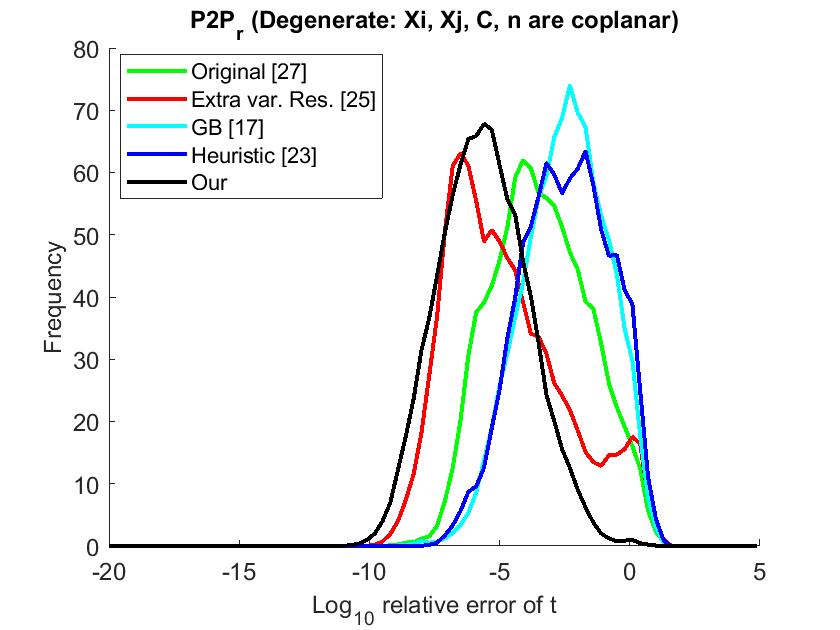}\\
 \includegraphics[width=0.24\linewidth]{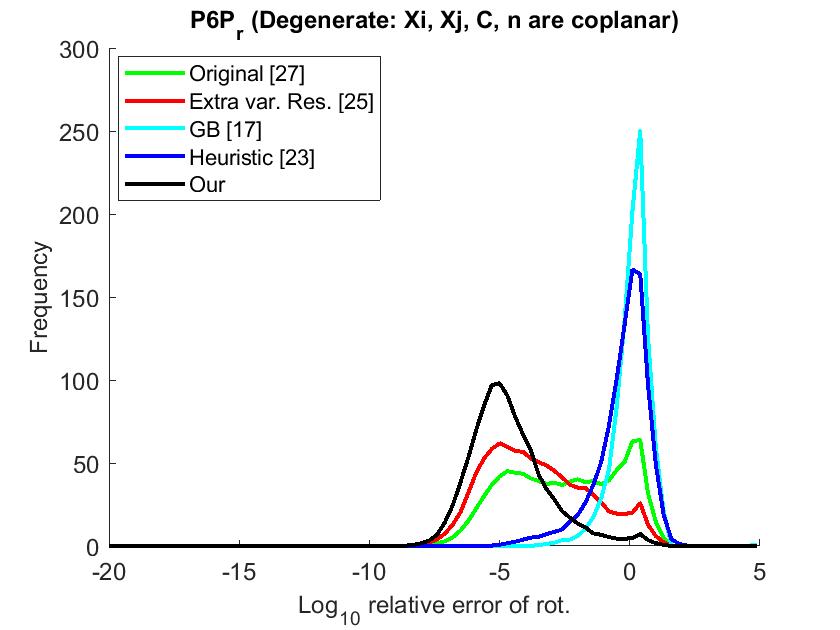}
\includegraphics[width=0.24\linewidth]{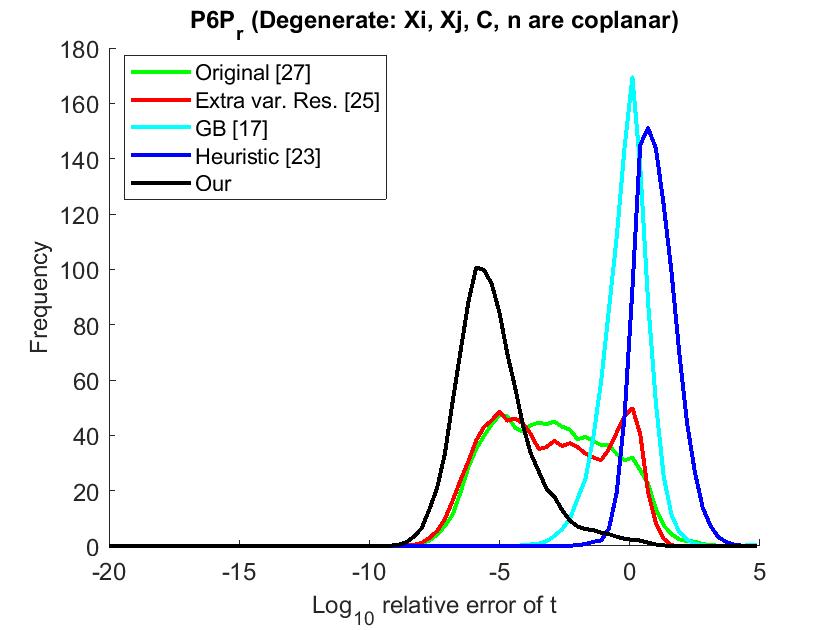}
\includegraphics[width=0.24\linewidth]{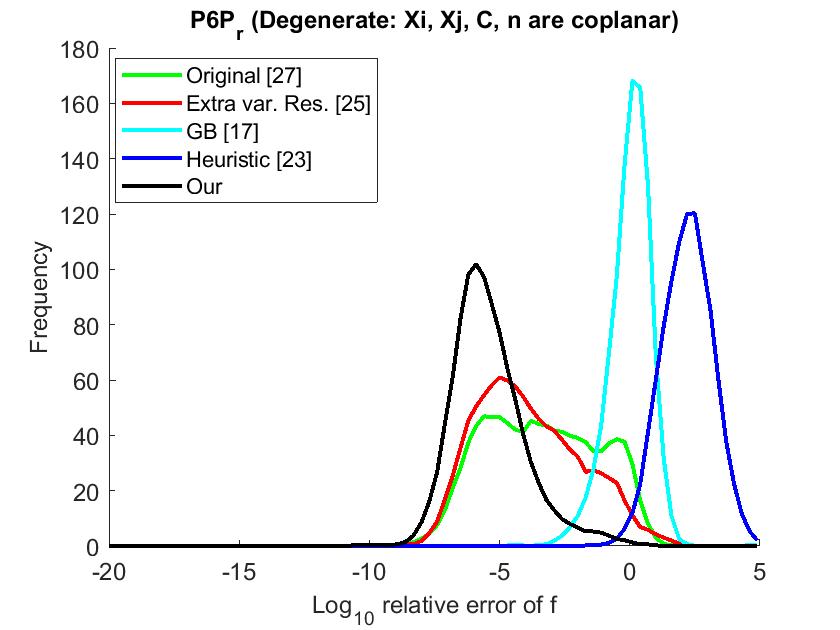}
\caption{Top: Relative errors of (a) rotation and (b) translation for the \PfP problem, (c) rotation and (d) translation for the \PtP problem measured w.r.t. ground truth on 1K synthetic degenerate scenes. Bottom: Relative errors of (e) rotation, (f) translation and (g) focal length for the \PsP problem measured w.r.t. ground truth on 1K synthetic degenerate scenes. }
\label{fig:degenscenehistograms}
\vspace{-0.15cm}
\end{figure*}

\begin{table*}
\centering 
\scalebox{0.68}{
\begin{tabular}{|c|c|c|c|c|c||c|c|c|c|c||c|c|c|c|c|}
 \hline
Comp. step & \multicolumn{5}{|c||}{\PfP} & \multicolumn{5}{|c|}{\PsP} & \multicolumn{5}{|c|}{\PtP} \\ 
 \hline
 & Orig~\cite{haner2015absolute} & GB~\cite{larsson2017efficient} & \cite{DBLP:conf/cvpr/LarssonOAWKP18} & Res~\cite{bhayani2019sparse} & \bf{Our} &  Orig~\cite{haner2015absolute} & GB~\cite{larsson2017efficient} & \cite{DBLP:conf/cvpr/LarssonOAWKP18} &  Res~\cite{bhayani2019sparse} & \bf{Our} & Orig~\cite{haner2015absolute} & GB~\cite{larsson2017efficient} & \cite{DBLP:conf/cvpr/LarssonOAWKP18} & Res~\cite{bhayani2019sparse} & \bf{Our} \\
 \hline
G-J/QR & $280 \times 399$ & $199 \times 215$ & $199 \times 215$ & $78 \times 93$ & -  & $648 \times 917$ & $636 \times 654$ & $398 \times 416$ & $248 \times 300$ & - & -  & $913\times 937$ & $597 \times 621$ & $142 \times 174$ & - \\
 \hline
EIG & $44 \times 44$ & $16\times 16$ & $16\times 16$ & $25 \times 25$ & - & $41 \times 41$ & $18 \times 18$ & $18 \times 18$  & $52 \times 52$ & -  & - & $24 \times 24$ & $24 \times 24$ & $32 \times 32$ & - \\
\hline
GEP & - & - & - & - & $36\times36$  & - & -  &  - & - & $110 \times 110$ &  $160 \times 160$ &-  &- &- & $124 \times 124$ \\
\hline
\hline
Time(ms) & - & $0.4937$  & $0.6683$ & $0.3344$ & $0.4743$ & - & $4.6193$ & $2.0822$ & $1.53$  & $5.192$ & - & $10.5689$  & $4.9556$ & $0.7292$  & $6.5612$  \\
\hline
\end{tabular}}
\caption{Comparison of important computation steps and running times of different solvers for three studied problems.}
\label{tb:sizecomparison}
\vspace{-0.7cm}
\end{table*}

\subsection{Solutions}
\vspace{-0.2cm}
\PAR{\PfP}: The absolute pose problem for calibrated cameras is originally solved by Haner \etal in \cite{haner2015absolute}, based on the \gb method. The elimination template matrix has size $280 \times 399$. This problem has only 16 solutions. However, in \cite{haner2015absolute} the authors create a larger action matrix of size $44 \times 44$ and use a basis selection method based on column-pivoting QR factorization~\cite{byrod2007improving} to improve the numerical stability of the solution. 
For generating solvers by using the state-of-the-art approaches as well as our proposed method, we reduce the input polynomials by performing symbolic Gaussian elimination in the offline stage. Using the standard \gb based approach~\cite{larsson2017efficient} we achieved an elimination template of size $199 \times 215$ and an eigenvalue formulation of size $16 \times 16$. In this case the heuristic-based \gb method presented in~\cite{DBLP:conf/cvpr/LarssonOAWKP18} did not improve the size of the final elimination template matrix. The \rs-based approach~\cite{bhayani2019sparse} solves this problem by computing a matrix inverse of size $78 \times 78$, followed by an eigendecomposition of size $25 \times 25$.
Canny-Emiris~\cite{DBLP:journals/jacm/CannyE00} algorithm solves this problem by obtaining a monomial basis of size $400$. Heikkil{\"{a}}'s algorithm~\cite{DBLP:conf/iccv/Heikkila17} generates a monomial basis of size $35$ but the matrix $\M{M^\prime}(x_n)$ is rank deficient, $\text{rank}(\M{M^\prime}(x_n)) = 32$. 
This means we can not generate a correct solver using~\cite{DBLP:conf/iccv/Heikkila17} and hence, we omit this solver in experiments. Using our method we are able to generate a monomial basis $\V{x}^\prime$ of size $40$ resulting in a GEP,~\eqref{eq:GEP} of size $80 \times 80$. After removing $0$-eigenvalues using the method presented in Section~\ref{subsec:zeigrem} we have a GEP of size $36 \times 36$. 
\vspace{-0.1cm}
\PAR{\PsP}: For cameras with unknown focal length the original \gb solution~\cite{haner2015absolute} results in an elimination template matrix of size $648 \times 917$. After removing symmetries using the method presented in~\cite{larsson2016uncovering} the problem has only $18$ solutions. In \cite{haner2015absolute} the authors again create a larger action matrix of size $41 \times 41$, and used the basis selection QR algorithm~\cite{byrod2007improving} to improve the numerical stability of the solution. 
For generating solvers based on the other state-of-the-art approaches as well as our proposed method, we reduce input polynomials by performing symbolic Gaussian elimination in the offline stage. The \gb solution presented in~\cite{larsson2017efficient} results in a template matrix of size $636 \times 654$ and the action matrix of size $18 \times 18$. Using the heuristic based approach presented in \cite{DBLP:conf/cvpr/LarssonOAWKP18} we achieved a smaller template matrix of size $398 \times 416$. Unfortunately, the Canny-Emiris algorithm in this case does not terminate in a reasonable time. Heikkil{\"{a}}'s algorithm returns a monomial basis of size $82$ but the matrix $\M{M^\prime}(x_n)$ is rank deficient, with rank $76$. The extra variable \rs based approach~\cite{bhayani2019sparse} results in a solver that includes matrix inverse of size $248 \times 248$ and an eigendecomposition of size $52 \times 52$.

Using our method we are able to generate the monomial basis $\V{x}^\prime$ of size $145 \times 145$ resulting in a GEP~\eqref{eq:GEP} of size $290 \times 290$. After removing zero eigenvalues using the method presented in Section~\ref{subsec:zeigrem} we get a GEP of size $110 \times 110$. Note that in our method we do not remove symmetries that appear in the equations, while the solvers~\cite{haner2015absolute} and~\cite{larsson2017efficient} do remove the symmetries and simplify the solver. Our method can be, however, easily combined with the symmetry removal method~\cite{larsson2016uncovering} and we hope that such symmetry removal will improve our solution even further. 
\vspace{-0.1cm}
\PAR{\PtP}: This problem results in system of $6$ equations in $4$ unknowns. In \cite{haner2015absolute} the authors have tried to solve this problem with $4$ equations using the \gb method, however the final solver contains thousands of polynomials and is extremely slow. Our attempt to use the \gb automatic generator from~\cite{larsson2017efficient} leads to a huge solver with an elimination template of size $913 \times 937$ and an eigenvalue problem of size $24 \times 24$, which is further improved by the heursitic based approach~\cite{DBLP:conf/cvpr/LarssonOAWKP18} leading to a template matrix of size $597 \times 621$. In \cite{haner2015absolute}, the authors present a hidden-variable solution to this problem. This solution results in a GEP of size $160 \times 160$. The resultant based solver~\cite{bhayani2019sparse} leads to a matrix inverse of size $142 \times 142$ and an eigenvalue problem of size $32 \times 32$. Unfortunately, the Canny-Emiris algorithm in this case does not terminate in a reasonable time and we cannot generate a solver for it. We have tried Heikkil{\"{a}}'s algorithm for this problem, but as this problem has more equations than unknowns, we are not able to test for larger polynomial combinations, while smaller combinations leads to incorrect rank deficient solvers. Our algorithm tests for all polynomial combinations returning a resultant of size $160 \times 160$, then reduces to a GEP of size $124 \times 124$ after removing $0$-eigenvalues. 
\vspace{-1.2mm}
\subsection{Evaluation}
\noindent The performance of our hidden variable \rs based solver (generated using MATLAB) to these three studied problems is compared with solvers based on \gb~\cite{larsson2017efficient}, the heuristic based approach~\cite{DBLP:conf/cvpr/LarssonOAWKP18}, the \rs based approach of adding an extra polynomial~\cite{bhayani2019sparse} and the original approach~\cite{haner2015absolute}.
\subsection{Synthetic scenes}
To carry out the experiments, we have set up a synthetic scene for above mentioned problems with a camera of feasible focal length. We also consider a medium with reasonable refractive index in front of the camera. The image points are sampled from an image of reasonable size while the scene points were selected in a cube of approximate dimensions $\left[-100, 100\right]$ in each direction in front of the image plane. The setup ensures that the scene points are on the other side of the refractive medium as compared to the camera. 

\subsection{Degenerate scene configuration}
\noindent We also conducted experiments on scenes with ``almost'' degenerate configurations. For all the three problems, we generate a possible solver degeneracy by considering a setup where the scene points, the camera center $\V{C}$ and the normal vector $\vec{\V{n}}$ of the refractive plane ``almost'' lie in the same plane. For this purpose we randomly select a plane through $\V{C}$ and $\vec{\V{n}}$ and choose scene points which are very close to this plane at reasonable distance on the other side of the refractive plane. We refer to~\cite{haner2015absolute} for more details about various degenerate configurations for each of the three minimal problems.

The goal of the first experiment is to test the numerical stability of proposed solvers and compare them with the state-of-the-art solutions. Therefore, we test our solution only on noise free correspondences. Since all solvers are solving the same formulation of the problem, the performance on the noisy measurements and real data would be the same up to some numerical instabilities that already appear in a noise-less case. For performance of these solvers in real applications we refer the reader to~\cite{haner2015absolute}.
As each solver returns multiple solutions, we use the solution closest to the ground truth when measuring the numerical stability of the solvers. Graphs (a) and (b) in Figure~\ref{fig:randscenehistograms} show the distribution of relative rotation and translation error for the \PfP problem as computed by solvers based on all of the five considered approaches, while graphs (c) and (d) show the error distribution for rotation and translation for the \PtP problem. At the same time, graphs (e), (f) and (g) show the error distribution in ground truth values in rotation, translation and focal length for the \PsP problem. Similarly, Figure~\ref{fig:degenscenehistograms} shows ground truth errors for all the three problems for 1K close-to-degenerate scenes. We note from these graphs that our proposed approach achieves comparable stability for random scene configurations and significantly outperforms the state-of-the-art solvers for close-to-degenerate scene configurations.
For the sake of time comparison, we consider only the major computation steps performed by our solvers and the fastest available solvers for each of the studied problems. This is done to have a reasonably fair comparison of execution times for all three problems. The timing comparison, averaged over 1K runs of solvers on synthetically generated scenes, is shown in Table~\ref{tb:sizecomparison} for three most important computation steps. The MATLAB solvers were run on a standard 3.9 GHz i7 based computer.
As we have noted earlier, our approach relies on sacrificing the solver size to improve the stability. Therefore, solvers based on our proposed method for three considered problems are slower than the fastest ones based on the extra variable resultant-based method~\cite{bhayani2019sparse}. However, the new solvers are significantly numerically more stable than the ones based on~\cite{bhayani2019sparse}.

\section{Conclusions}
\noindent 
 In this paper we explored an approach of hiding a variable in order to compute sparse resultant based solver for minimal problems where we solve a larger eigenvalue problem and eliminate the need to perform matrix inverse, thus leading to improved solver stability. Our approach 
also includes improvements to extend previous hidden variable sparse resultant based methods~\cite{DBLP:conf/iccv/Heikkila17, DBLP:journals/jacm/CannyE00} in order to generate stable solvers of reasonable size. This method can be easily automated and it moves most of the computation to the pre-processing step. We demonstrated the stability of our solver on absolute pose estimation problems and tested on synthetic scenes for random as well as almost degenerate configurations. Apart from the solvers generated based on the original approaches~\cite{haner2015absolute}, we also compare the stability of our proposed solvers with those computed based on \gb~\cite{larsson2017efficient,DBLP:conf/cvpr/LarssonOAWKP18} and the extra variable \rs approach~\cite{bhayani2019sparse} where our solver outperforms the state-of-the-art solvers in terms of stability.
The proposed method has a potential to improve numerical stability of other minimal solvers especially for close-to-degenerate configurations. \vspace{-0.2cm}
\section{Acknowledgement}
The authors would like to thank Academy of Finland for the financial support of this research (grant no. 297732). Zuzana Kukelova was supported by OP VVV project Research Center for Informatics reg. no. CZ.02.1.01/0.0/0.0/16\_019/0000765 
\bibliographystyle{IEEEtran}
\bibliography{egbib}

\end{document}